\documentclass[runningheads]{llncs}
\usepackage[T1]{fontenc}
\usepackage{graphicx}
\usepackage{multirow}
\usepackage{array}
\usepackage{color}
\usepackage{algorithm}
\usepackage[noend]{algpseudocode} 
\algdef{SE}[DOWHILE]{Do}{doWhile}{\algorithmicdo}[1]{\algorithmicwhile\ #1}%
\algnewcommand{\IIf}[1]{\State\algorithmicif\ #1\ \algorithmicthen}
\algnewcommand{\IElse}[1]{\State \algorithmicelse\ #1 }
\algnewcommand{\EndIIf}{\unskip\ 
}
\usepackage{wrapfig}
\usepackage{float}
\usepackage{bm}
\usepackage{amssymb}
\usepackage{amsmath}
\usepackage{comment}
\usepackage{caption}
\usepackage{subcaption}
\usepackage{graphicx}
\usepackage{multirow}
\usepackage{colortbl}
\usepackage{xspace}
\usepackage{booktabs}
\usepackage{stmaryrd}
\usepackage{paralist}
\usepackage{cite}
\usepackage[normalem]{ulem}

\usepackage{eso-pic}
\usepackage[breaklinks]{hyperref}
\hypersetup{
colorlinks = true, 
linkcolor=black, 
citecolor=black, 
urlcolor=black 
}
\DeclareCaptionFont{f7}{\fontsize{7}{8}\selectfont}
\usepackage{lineno}

\usepackage{subcaption}
\newcommand{\tool
}{\textsc{BMiner}\xspace}

\newcommand{\toolb
}{\ensuremath{\textsc{BMiner}_{\mathsf{E}}}\xspace}

\newcommand{\toolg
}{\ensuremath{\textsc{BMiner}_{\mathsf{G}}}\xspace}

\definecolor{mygreen}{RGB}{200, 255, 200}
\newcommand{\tbgreen}{\cellcolor{mygreen}}

\definecolor{myyellow}{RGB}{255, 245, 200}

\definecolor{myred}{RGB}{255, 220, 220}
\newcommand{\tbred}{\cellcolor{myred}}

\newcommand{\argmax}{\mathop{\rm arg~max}\limits}

\newcommand{\R}{\ensuremath{\mathbb{R}}\xspace}
\newcommand{\N}{\mathbb{N}}
\newcommand{\Npos}{\N_{\ge 0}}
\newcommand{\inputSpec}{\ensuremath{\mathrm{\Phi}}\xspace}
\newcommand{\outputSpec}{\ensuremath{\mathrm{\Psi}}\xspace}
\newcommand{\reluSpec}{\ensuremath{\mathrm{\Gamma}}\xspace}
\newcommand{\reluAction}{\ensuremath{a}\xspace}

\newcommand{\reluActionPos}{r^+}
\newcommand{\reluActionNeg}{r^-}
\newcommand{\reluHeuristic}{\mathsf{H}}
\newcommand{\true}{\ensuremath{\mathsf{true}}\xspace}
\newcommand{\false}{\ensuremath{\mathsf{false}}\xspace}

\newcommand{\specDist}{\hat{p}}

\newcommand{\queue}{\ensuremath{\mathrm{Q}}\xspace}
\newcommand{\bab}{\ensuremath{\mathtt{BaB}}\xspace}
\newcommand{\dnn}{\ensuremath{N}\xspace}
\newcommand{\verifier}{\ensuremath{\mathtt{AppVer}}\xspace}
\newcommand{\istart}{\ensuremath{\mathit{i_l}}\xspace}
\newcommand{\iend}{\ensuremath{\mathit{i_r}}\xspace}

\newcommand{\relu}{\ensuremath{\mathsf{ReLU}}\xspace}
\newcommand{\reluInput}{\hat{x}}

\newcommand{\specTree}{\ensuremath{\mathcal{T}}\xspace}
\newcommand{\overAppro}{\ensuremath{\hat{\Omega}}\xspace}
\newcommand{\pathNum}{\ensuremath{n}\xspace}
\newcommand{\boundary}{\ensuremath{b}\xspace}

\newcommand{\abcrown}{{$\alpha\beta$-\textsf{Crown}}\xspace}
\newcommand{\mnist}{\texttt{MNIST}\xspace}
\newcommand{\cifar}{\texttt{CIFAR-10}\xspace}

\newcommand{\OVAL}{\texttt{OVAL21}\xspace}
\newcommand{\base}{\texttt{BASE}}
\newcommand{\deep}{\texttt{DEEP}}
\newcommand{\wide}{\texttt{WIDE}}
\newcommand{\lfour}{\texttt{L4}}
\newcommand{\ltwo}{\texttt{L2}}

\newcommand{\oliva}{{\textsf{Oliva}}\xspace}
\newcommand{\planet}{\textsf{Planet}\xspace}

\newcommand{\mono}{\texttt{mono}}

\newcommand{\myparagraph}[1]{\smallskip\noindent{\bf #1.}}

\spnewtheorem{mytheorem}{Theorem}
{\bfseries}{\rmfamily} 
\spnewtheorem{mylemma}{Lemma}
{\bfseries}{\rmfamily} 
\spnewtheorem{myexample}{Example}{\bfseries}{\rmfamily}
\spnewtheorem{myassumption}[mytheorem]{Assumption}{\bfseries}{\rmfamily}
\spnewtheorem{mydefinition}{Definition}{\bfseries}{\rmfamily}
\spnewtheorem{myremark}{Remark}{\bfseries}{\rmfamily}

\newcounter{researchquestionCount}
\newcommand{\researchquestion}[1]{\stepcounter{researchquestionCount}\medskip\noindent\parbox{0.97\textwidth}{{\bf RQ\arabic{researchquestionCount}} {\it #1}}\smallskip}

\def\orcidID#1{\kern .08em\href{https://orcid.org/#1}{\includegraphics[keepaspectratio,width=0.9em]{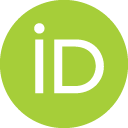}}}

\begin{document}
%
\title{Mining Verdict Boundaries for Neural Network Verification}

%
\author{
Jiawei Ren\inst{1}\orcidID{0000-0001-7635-468X} \and
Guanqin Zhang\inst{1}\orcidID{0000-0002-3844-8180} \and
Zhenya Zhang\inst{2,3}\orcidID{0000-0002-3854-9846} \and
Yulei Sui\inst{1} \orcidID{0000-0002-9510-6574}
}

\authorrunning{J. Ren et al.}

\institute{University of New South Wales, Sydney, Australia\\
\email{\{jiawei.ren, guanqin.zhang, y.sui\}@unsw.edu.au} 
\and
Kyushu University, Fukuoka, Japan\\
\email{zhang@ait.kyushu-u.ac.jp} 
\and National Institute of Informatics, Tokyo, Japan
}
\AddToShipoutPictureFG*{%
  \AtPageLowerLeft{%
    \put(\LenToUnit{0.09\paperwidth},\LenToUnit{11mm}){%
      \parbox{0.82\paperwidth}{%
        \centering
        \fontsize{6.8}{8}\selectfont
        This version of the contribution has been accepted for publication,
        after peer review, but is not the Version of Record and does not
        reflect post-acceptance improvements or corrections.
        The Version of Record is available online at:
        \href{https://doi.org/10.1007/978-3-032-26204-2_4}
        {\nolinkurl{https://doi.org/10.1007/978-3-032-26204-2_4}}.%
      }%
    }%
  }%
}
\maketitle              

\vspace{-1.5em}

\begin{abstract}
Branch and Bound (\bab) aims to achieve complete verification of neural networks by adaptively partitioning the problem and applying off-the-shelf verifiers to subproblems. 
Its problem-splitting history can be represented as a tree, where each subproblem corresponds to a child node. A key problem of \bab lies in searching for the \emph{verdict boundaries} across all the paths that divide the verified and unverified subproblems.
We observe that the existing \bab approach tackles this problem by solving each expensive subproblem sequentially along the tree path as its depth increases, requiring costly bounds propagation 
at every visited \bab tree node (i.e., subproblem), which is inefficient.
To address this issue, we propose effective search approaches that leverage the monotonicity of each path to efficiently and precisely locate the verdict boundary by simultaneously splitting multiple activation functions (e.g., ReLU), rather than processing them one at a time as in the classical approach.
Our approach performs an effective exponential search along each path, allowing us to skip many boundary-unrelated subproblems when identifying the verdict boundary. The enhanced version further improves this process by estimating the boundary’s position using quantitative information obtained from subproblem solving.
We perform experimental evaluation on commonly-used benchmarks to assess our proposed techniques, and compare them with recent \bab-based approaches.
\end{abstract}
\section{Introduction}

The rapid adoption of neural networks in safety-critical settings, such as autonomous driving, makes unexpected model failures potentially severe. Their susceptibility to adversarial perturbations~\cite{madry2017towards, goodfellow2015explaining} underscores the need for rigorous assurance. Neural network verification offers provable guarantees of safety and robustness under specified input conditions~\cite{lukina2021into, henzinger2020outside,mandal2024formally} and has therefore become a rapidly growing area within formal verification research.

\myparagraph{Existing Efforts}
Typical neural network verification approaches include completeness-driven constraint solving with a timeout budget~\cite{katz2017reluplex,tjeng2018evaluating} and approximation-based methods~\cite{singh2019abstract,singh2018deepz,wang2018efficient}. 
While the former aims to compute exact bounds, the latter trades precision for efficiency. 
To mitigate the incompleteness of approximation-based methods, \emph{Branch and Bound} (\bab)~\cite{bunel2020branch} has emerged as the state-of-the-art abstraction refinement approach.
Given a specification, \bab employs an efficient approximation-based verifier to compute a distance $\specDist$ that signifies \emph{how much} the network satisfies the specification; if $\specDist$ is negative, it splits the problem into subproblems (e.g., on input intervals~\cite{wang2018formal} or ReLU phases~\cite{bunel2020lagrangian}). It repeats this process on each subproblem, which can yield a tighter approximation and hence a greater $\specDist$, for each subproblem, until all the subproblems are verified (i.e., $\specDist > 0$)~\cite{wang2021beta}.

\myparagraph{Motivation and Insights}
In \bab, the iterative problem splitting process can be structured as a tree, where each node identifies a subproblem. The objective is to confirm that each tree path ends with a verified node, which is a necessary condition to verify the original problem. Moreover, since problem splitting reduces approximation error, $\specDist$, which measures how much the over-approximation satisfies the specification ($\hat{p} > 0$ implies satisfaction), often exhibits a high degree of monotonicity along each path as the node depth increases. We empirically confirm this behavior using a state-of-the-art \bab approach~\cite{wang2021beta}. Because of this, there must exist a \emph{boundary point} (regardless of special cases, such as when the problem is verified at the root) in each path that divides the verified nodes and the unverified nodes, which require no further problem splitting. \bab essentially needs to find such a verdict boundary for each path of the tree.
 
However, existing \bab approaches, e.g.,~\cite{bunel2020branch,wang2021beta}, identify the verdict boundary in each path by calling an approximated verifier on each node \emph{one by one} along the path, to locate the boundary point. 
Since each visit of a node involves a non-trivial problem-solving process by approximated verifiers, this introduces significant time costs. 
On the other hand, due to the monotonicity of $\specDist$ in each path, the problem of finding the boundary point in each path boils down to a search problem for a value in a nearly sorted array. 
To this end, we can devise more efficient approaches to find the verdict boundary. 

\myparagraph{Our Solution} In this paper, we propose a new verification approach, \tool, which performs efficient per-path boundary mining within the \bab tree. 
By exploiting the monotonicity of $\specDist$ in each tree path, \tool formulates the location of the verdict boundary as a search problem in a nearly sorted array and thereby devises two efficient search approaches that, instead of visiting each tree node one-by-one, skip significant node visits in each tree path.

The first approach \toolb adapts the classic \emph{exponential search}~\cite{bentley1976almost}. Given a path, \toolb involves two phases to locate the boundary point: first, it identifies a range that contains the verdict boundary point, by progressively sliding and expanding the range with an exponentially increasing size; after fixing such a range, it locates the exact boundary point position by binary search. 

While \toolb can reduce the number of node visits compared to classic \bab, it may still need too many node visits to identify the range that contains the boundary point in a path. 
To further improve this, we devise \toolg, which leverages the quantitative $\specDist$ of nodes to estimate its \emph{gradient} (more precisely, its changing rate) in a path, by which it predicts the position of the boundary point. While the prediction may not be precise, by checking the node at the predicted position, we can efficiently update the gradient information and continue our search until we find the boundary point. 

We compare with two state-of-the-art tools as our baselines, on 500 verification problem instances from the widely used \mnist and \cifar datasets. The evaluation results demonstrate that our proposed approach can reduce the average verification time cost by 17\% -- 30\%, with a maximum reduction of 44.7\% across five neural networks.

\vspace{-1mm}
\section{Preliminaries}\label{sec:preliminary}
\vspace{-1mm}

\subsection{Neural Network Verification Problem}
We define feed-forward neural networks following standard literature $\text{\cite{katz2017reluplex,singh2019abstract,bunel2020branch}}$.

\begin{mydefinition}[Neural networks]\label{def:neuralNetwork}
A (feed-forward) neural network 
$\dnn:\mathbb{R}^n\!\to\!\mathbb{R}^m$ 
consists of alternating affine transformations and nonlinear activations $\sigma$.
Let $\bm{x}_0$ be the input.  
For each layer $i$ ($i\in\{1, \ldots, L\}$), 
\(
    \bm{x}_i = \sigma(W_i \bm{x}_{i-1} + B_i),
\)
where $W_i,B_i$ are the weights and biases.  
The dimension of $\bm{x}_i$ equals the number of neurons in layer $i$.
Following established works~\cite{liu2021algorithms}, we adopt ReLU in our neural networks, defined as $\relu(x) = \max(0, x)$.
\end{mydefinition}

Specifications are used to formalize desired properties of neural networks. We adopt the notation in Def.~\ref{def:spec} as our specification formalism, which is sufficiently expressive for commonly-used properties, such as \emph{local robustness}~\cite{goodfellow2015explaining}.
\begin{mydefinition}[Specification]\label{def:spec}
    We define a \emph{specification} for a neural network as a pair $(\inputSpec, \outputSpec)$ that includes an input specification $\inputSpec$ and an output specification $\outputSpec$. 
    The \emph{input specification} $\inputSpec$ is a predicate over the input region, and the output specification $\outputSpec$ is a predicate over the output of the neural network.
    Specifically, we represent the output specification $\outputSpec$ as $f(\dnn(\bm{x})) > 0$, where $f \colon \R^m \to \R$ is a function that maps an $m$-dimensional vector to a real number.
\end{mydefinition}

\begin{mydefinition}[Verification problem]\label{def:vp}
    Given a neural network $\dnn$ and a specification $(\inputSpec, \outputSpec)$, a verification problem aims to answer the question whether $\outputSpec(\dnn(\bm{x}))$ holds, for any input $\bm{x}$ that holds $\inputSpec(\bm{x})$. A verifier is a tool designed to solve this problem. It either returns $\true$, certifying that $\dnn$ satisfies the specification, or returns $\false$ with a counterexample.
\end{mydefinition}

We now demonstrate how our notations formalize a verification problem as in Def.~\ref{def:vp} against local robustness, a property frequently studied in image classification. Local robustness requires that a neural network classifier consistently assigns the same label to two images, $\bm{x}$ and $\bm{x}_0$, where $\bm{x}$ is derived from $\bm{x}_0$ by applying small perturbations. 
Formally, given a reference input $\bm{x}_0$, the input specification $\inputSpec(\bm{x})$ enforces that $\bm{x}$ should hold that $\|\bm{x} - \bm{x}_0\|_\infty \leq \epsilon$, where $\|\cdot\|_\infty$ is the $\ell^\infty$-norm distance metric, and $\epsilon$ is a small positive real. 
The output specification $\outputSpec(\dnn(\bm{x}))$ requires that $\min_{1\le i \le m, i\neq i_0} (\dnn(\bm{x})_{i_0} - \dnn(\bm{x})_i) > 0$ where $\dnn(\bm{x})_i$ denotes the $i$-th component of the output vector $\dnn(\bm{x})$, and $i_0$ is the label assigned to $\bm{x}_0$ by the neural network, i.e., $ i_0 = \argmax\textstyle_{1 \leq i \leq m}\dnn(\bm{x}_0)_i$.

\subsection{Verification by Branch-and-Bound (\bab)}\label{sec:bab}

\begin{wrapfigure}[23]{r}{0.49\linewidth}
    \begin{minipage}{\linewidth}
\vspace{-4.4em}
\begin{algorithm}[H]
\caption{Branch and Bound~\cite{bunel2020branch}}
\label{alg:bab}
\footnotesize
\begin{algorithmic}[1]
\Require A neural network $\dnn$, an input specification $\inputSpec$, an output specification $\outputSpec$, an approximated verifier $\verifier(\cdot)$, and a ReLU selection heuristic $\reluHeuristic(\cdot)$. 
\Ensure A $\mathit{verdict}\in\{\true, \false\}$.
\State $\queue \gets \{\top\}$ \label{line:initQueue}, $\specTree\gets \emptyset$ \label{line:initialSpecTree} 
\State \Call{BaB}{$\queue$}\label{line:startBaB} 
\Function{BaB}{$\queue$}
\label{line:functionBaB}
\If{$\Call{Empty}{\queue}$}\label{line:judgeQEmpty}
\State \Return \true \label{line:Qempty} 
\EndIf
\State $\reluSpec \gets \Call{Pop}{\queue}$ \label{line:popQ} 
\State $\specDist \gets \verifier(\dnn, \inputSpec, \outputSpec, \reluSpec)$ \label{line:callAppVerifier} 
\State $\specTree\gets\specTree\cup \{\langle\reluSpec, \specDist\rangle\}$ \label{line:treeRecord} 
\If{$\specDist < 0$} 
\If{$|\reluSpec| = K$} \label{line:validCE}
\State \Return $\false$ \label{line:babFalse}
\Else \label{line:invalidCE}
\State $\{\reluActionPos_i, \reluActionNeg_i\}\gets \reluHeuristic(\reluSpec)$\label{line:selectNextRelu} 
\For{$a \in \{\reluActionPos_i, \reluActionNeg_i\}$}
\State $\queue \gets \queue\cup \{\reluSpec\land a\}$ 
\label{line:push} 
\EndFor
\EndIf
\EndIf
\State  \Return $\Call{BaB}{\queue}$ \label{line:recursiveCallBaB} 
\EndFunction
\end{algorithmic}
\end{algorithm}
    \end{minipage}
\end{wrapfigure}
\bab~\cite{bunel2020branch} serves as the backbone for several advanced verification tools, such as \abcrown~\cite{wang2021beta}. 
Essentially, it is a \emph{``divide-and-conquer''} strategy that adaptively splits a verification problem into its subproblems and leverages off-the-shelf approximated verifiers to tackle the resulting subproblems. Since approximated verifiers can yield tighter over-approximations on subproblems, \bab mitigates their completeness limitations when applied directly to the original problem.

\begin{wrapfigure}[8]{r}{0.34\textwidth}
    \centering
    \vspace{-2em}
    \includegraphics[width=0.3\textwidth]{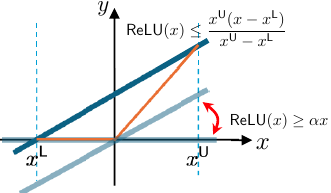}
    \vspace{-1em}
    \caption{Linear relaxation of ReLU function.}
    \label{fig:deeppoly}
\end{wrapfigure}

\myparagraph{Approximated verifiers} Various techniques have been proposed to construct over-approximation, e.g., linear relaxation by \emph{DeepPoly}~\cite{singh2019abstract}  that uses linear constraints to bound each ReLU's output (see Fig.~\ref{fig:deeppoly}).
Notably, \bab is orthogonal to these approaches, i.e., it works with any of the selected ones. For a verification problem, an approximated verifier \verifier returns a quantity $\specDist\in\R$ called a \emph{verifier assessment}~\cite{zhang_ecoop}, computed by $\specDist = \min_{\bm{y}\in \overAppro}f(\bm{y})$, where $\overAppro$ is the over-approximated region of the neural network output, and $f$ is the function in output specification as defined in Def.~\ref{def:spec}. 
Here $\specDist$ indicates \emph{how much} the over-approximation satisfies the specification: If $\specDist$ is positive, the original output satisfies the specification, thereby certifying the neural network.

\begin{mydefinition}[ReLU specification]\label{def:reluSpec} 
Let \dnn be a neural network with $K$ neurons, and $\reluInput_i\in\R$ ($i\in\{1, \ldots, K\}$) be the input for the ReLU function in the $i$-th neuron. 
Introduce two literals 
\(\reluActionPos_i := (\reluInput_i \ge 0),\ 
\reluActionNeg_i := (\reluInput_i < 0)
\).
  A \emph{ReLU specification} is a conjunction of literals that fixes the activation sign of a (possibly empty) set of distinct neurons: $\;\reluSpec =
    \bigwedge_{i\in I}
      \ell_i$,
  where $I\subseteq\{1,\dots ,K\}$ and
  $\; \ell_i \in\{\,\reluActionPos_i,\,\reluActionNeg_i\}$.
The length of the specification is defined as $|\reluSpec| = |I|$.  
When $I=\varnothing$, we write $\reluSpec = \top$ and call it an empty specification.
\end{mydefinition}

Given a neuron $i$, w.r.t the sign of its pre-activation value, $\reluActionPos_i$ and $\reluActionNeg_i$ can be derived to split the ReLU function into two linear functions. 
By adding each proposition as an additional constraint, the verification problem can be divided into two subproblems. If both subproblems are verified, since $\reluActionPos_i$ and $\reluActionNeg_i$ cover all possible cases of the ReLU input, the original problem is also verified.

The \bab algorithm is presented in 
Alg.~\ref{alg:bab}. In addition to input/output specification, we allow \verifier to take an additional argument, i.e., a ReLU specification $\reluSpec$, which identifies a subproblem obtained by adding $\reluSpec$ (i.e., the conjunction of a number of propositions about ReLU inputs) as a constraint to the original problem. Alg.~\ref{alg:bab} uses a queue \queue to maintain the (sub)problems to be solved, which is initialized as a set that contains only $\top$, i.e., the original problem. The algorithm starts with a call of function \textsc{BaB} with the initial $\queue$ (Line~\ref{line:startBaB}).

\begin{compactenum}[i)]
    \item First, \bab applies \verifier to the original problem (Line~\ref{line:callAppVerifier}): if \verifier returns a positive $\specDist$, it implies that the original problem is verified and so the verification can be terminated with $\true$ returned; 
    \item \label{step:case} In the case the problem cannot be verified (i.e., $\specDist < 0$), it splits the problem into two subproblems. This is achieved by selecting a ReLU in the neural network according to a pre-defined ReLU selection heuristic $\reluHeuristic$ that returns two propositions $\{\reluActionPos_i, \reluActionNeg_i\}$ about the input of the selected ReLU, based on a given ReLU specification $\reluSpec$ (Line~\ref{line:selectNextRelu}; details are introduced later). The selected ReLU $i$ derives two subproblems, identified by $\reluSpec\land\reluActionPos_i$ and $\reluSpec\land\reluActionNeg_i$ respectively, which are pushed into $\queue$;
    \item Then, the function \textsc{BaB} is called recursively with the updated $\queue$ (Line~\ref{line:recursiveCallBaB}), so it can apply \verifier to the new subproblems. For each subproblem, it goes through a similar process with the original problem: the subproblem can be verified if $\specDist > 0$;  otherwise, it needs to be further split;
    \item The original problem (identified by $\top$) can be verified if all of the subproblems are verified, in which case \queue will be empty  (Line~\ref{line:Qempty}).
\end{compactenum}

\bab needs a ReLU selection heuristic $\reluHeuristic$ that orders neurons and selects the next ReLU according to the specification. Existing strategies such as DeepSplit~\cite{henriksen2021deepsplit} and FSB~\cite{de2021improved} can be integrated, as our method is orthogonal to them. Following an existing neural network verification approach~\cite{wang2021beta}, we adopt and extend the state-of-the-art ReLU selection strategy~\cite{de2021improved} (detailed in \S{}\ref{sec:evaluation}).

\vspace{-1mm}
\section{Verdict Boundary Insight and Formulation}
\label{sec:formulation}
\vspace{-1mm}

We introduce a formulation that models \bab-based verification as a search problem. 
Strictly, this formulation should be on the premise of the near-monotonicity of $\specDist$ when problem splitting happens, but this may not always hold for some recent implementations of \bab, such as \abcrown~\cite{wang2021beta}. Our empirical study shows that such non-monotonic splitting is rare, and even when it occurs, it is typically local and does not change the overall monotonic trend of $\specDist$; thus, it does not affect the effectiveness of our approach, as confirmed experimentally.

\subsection{Near-Monotonic Behaviour in BaB Tree Paths}\label{subsec:monotone}
The \bab approach (Alg.~\ref{alg:bab}) produces a binary tree, called the \bab tree.

\begin{mydefinition}[\bab tree]\label{def:babTree}
    In \bab, a binary tree $\specTree$, called a \bab tree, is used to record the (sub)problems produced during the execution of \bab. 
    Each node is denoted as a tuple $\langle\reluSpec, \specDist\rangle$, which consists of a ReLU specification $\reluSpec$ that identifies a (sub)problem, and a verifier assessment $\specDist$ that indicates the verification result of the (sub)problem by applying \verifier. 
\end{mydefinition}

In general, as a \emph{divide-and-conquer} approach, \bab aims to continuously refine the approximation to increase  $\specDist$, i.e., the distance between the over-approximated output region $\overAppro$ and the negation of the specification, via a sequence of problem splitting in different paths. In other words, during a sequence of problem splitting, $\specDist$ is expected to keep monotonically increasing until it becomes positive at some point, which signifies a successful verification of the subproblem.

However, problem splitting may not always produce an increased $\specDist$, depending on the adopted approximated verifier. 
For example, \bab with \planet~\cite{ehlers2017formal} as the verifier, holds such monotonicity, because every problem splitting can lead the over-approximation $\overAppro$ to a subset $\overAppro^\dagger \subseteq \overAppro$.
In recent approaches, e.g., \abcrown~\cite{wang2021beta}, this monotonicity may be broken. 
Instead of using a fixed linear bound for ReLU, they search for an optimal lower bound by varying $\alpha$ (i.e., the slope of the linear constraints in Fig.~\ref{fig:deeppoly}). 
This may lead to the situation where the over-approximation $\overAppro^\dagger$ after problem splitting is not necessarily a subset of $\overAppro$ before the splitting. 
Consequently, $\specDist$ may decrease. To understand how frequently this can happen, we perform an empirical study for \abcrown.


\begin{wrapfigure}[11]{r}{0.58\textwidth}
    \centering
    \vspace{-2.5em}
\includegraphics[width=\linewidth]{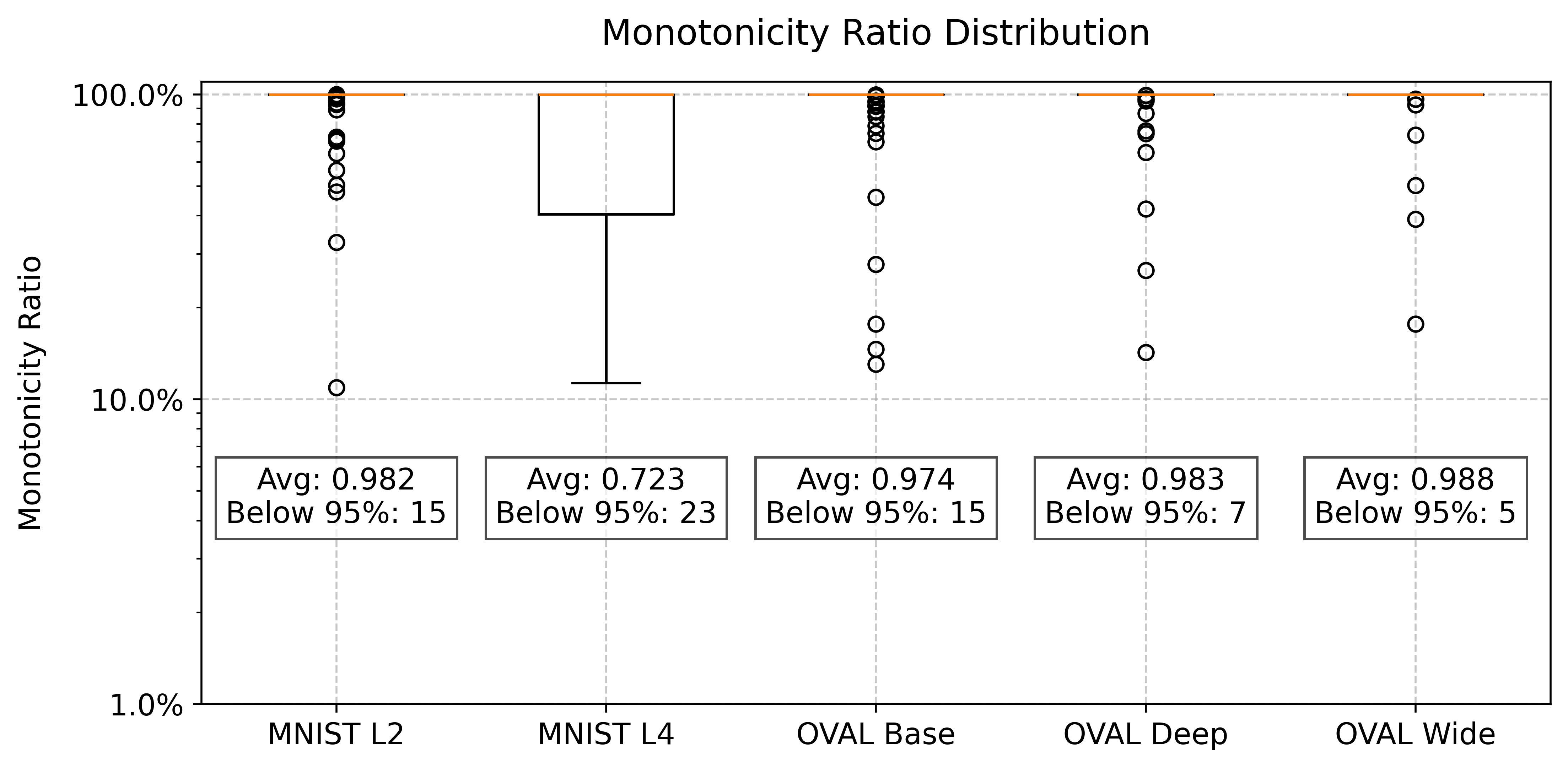}
\vspace{-1.5em}
    \caption{Distribution of ratio of monotone paths}
    \label{fig:mono_dis_Tree}
\end{wrapfigure}
\myparagraph{Empirical evidence on \abcrown}\label{study}
We adopt the same set of benchmarks used in our experimental evaluation (see Table~\ref{tab:benchmark}).
We apply \abcrown to solve each problem, and record the resulting \bab tree.

Fig.~\ref{fig:mono_dis_Tree} shows a box plot of the distribution of the ratio of the monotonic paths over all the paths in each \bab tree. By a monotonic path, we mean the paths in which $\specDist$ never decreases in a node against its parent. We annotate the relevant statistical data, including the average ratio and the number of trees whose ratio is below 95\%. By Fig.~\ref{fig:mono_dis_Tree}, we can find that, for many \bab trees, all their paths are monotonic, as evidenced by the very narrow distance between the lower and upper quartiles.  
Even if there are some trees in which non-monotonic paths exist, the number of such trees is not very large in most models. The notable exception is $\mnist_{{{\lfour}}}$, for which there is a considerable number of trees whose monotonic path ratio is relatively low. This could be attributed to our way of counting, i.e., as long as there exists a non-monotonic node, we count the path as non-monotonic; because of this, if a non-monotonic node occurs at a location close to the root, it affects all the paths in its sub-trees, so it can make the ratio very low.

Note that the metric used in Fig.~\ref{fig:mono_dis_Tree}, i.e., the ratio of monotonic paths, is relatively strong, because monotonicity of the whole path requires that there does not exist even a single pair of nodes that breaks the monotonicity. 
Therefore, the metric used in Fig.~\ref{fig:mono_dis_Tree} is a relatively strong one, and is subject to the positions of such non-monotonic pairs of tree nodes. 
In Table~\ref{tab:monotable}, we further analyze the impact of these pairs on the whole paths.

\begin{wraptable}[8]{r}{0.49\linewidth}
    \centering
    \footnotesize
    \vspace{-2.0em}
    \caption{Monotonicity Statistics.}
    \label{tab:monotable}
    \vspace{-1em}
    \resizebox{\linewidth}{!}{%
    \begin{tabular}{lcccc}
    \toprule
        Model & Tree size  & $\text{Path}^{\mono}$ & $\text{Pair}^{\mono}$ & $\text{Triple}^{\mono}$\\
        \midrule
        $\mnist_{{{\ltwo}}}$ &163&98.1\%&99.8\%&69.1\%  \\
        $\mnist_{{{\lfour}}}$  &2535&72.3\%&96.6\%&74.7\%  \\
        $\OVAL_{{\base}}$ &896&97.4\%&99.7\%&95.8\%  \\
        $\OVAL_{{\deep}}$&366&98.3\%&99.8\%&97.3\%  \\
        $\OVAL_{{\wide}}$&877&98.8\%&99.9\%&99.7\%  \\
    \bottomrule
    \end{tabular}}
\end{wraptable}
\begin{compactitem}[$\bullet$]
    \item First, the column ``$\text{Path}^\mono$'' shows the overall ratio of the monotonic paths over all the paths of all the trees in the tasks of each model. This confirms that most of the paths hold monotonicity;
    \item Second, the column ``$\text{Pair}^\mono$'' shows how many pairs are non-monotonic, over all the pairs in the trees. We find that, the ratios are all greater than those in ``$\text{Path}^\mono$'', showing that mostly the non-monotonicity of paths happens, due to some individual pairs in the paths; 
    \item Third, the column ``$\text{Triple}^\mono$'' records that, among the non-monotonic pairs in ``$\text{Pair}^\mono$'', how many of them hold that, $\specDist$ in the subsequent node after the pair is increased compared to the first node in the pair; namely, this metric implies that how many non-monotonic pairs only happen locally and do not have a significant impact on the whole paths.
\end{compactitem}

The results indicate that, while some \bab implementations may not hold monotonicity, such violations do not happen frequently. Moreover, the impact on the whole path mostly remains local.
By this, we proceed to our formulation that deems \bab verification as a search problem.

\subsection{Verdict Boundary}
\label{subsec:formulation}

We first introduce the notion of verdict boundary in Def.~\ref{def:boundary}. Later, we show the connection between this notion and the objective of \bab-based verification. 
\begin{mydefinition}[Verdict boundary]
\label{def:boundary}
    Given a \bab tree that has $\pathNum$ paths, a \emph{verdict boundary} is an $\pathNum$-tuple $\langle \boundary_1, \ldots, \boundary_{\pathNum} \rangle$, where each $\boundary_i\in\Npos$ ($i\in\{1,\ldots,\pathNum\}$) is called a \emph{boundary point} such that, for each node $\langle\reluSpec, \specDist\rangle$ in the $i$-th path, if $|\reluSpec| \ge \boundary_i$, then $\specDist \ge 0$; otherwise if $|\reluSpec| < \boundary_i$, then $\specDist < 0$. 
    Specifically, $\boundary_i = 0$ if the root node of the \bab tree is verified; $\boundary_i = K+1$ ($K$ is the total number of ReLUs) if none of the nodes is verified in the $i$-th path.
\end{mydefinition}

Intuitively, each element $\boundary_i$ in a verdict boundary is a position in a tree path that divides the nodes respectively decided as verified and unverified by approximated verifiers. As $\specDist$ is \emph{mostly} monotonic in each path, the nodes preceding $\boundary_i$ are most likely to be unverified, and the nodes succeeding (including) $\boundary_i$ are most likely to be verified. Example~\ref{ex:boundary} visualizes a verdict boundary in a \bab tree.

\begin{wrapfigure}[8]{r}{0.63\textwidth}
    \centering
    \vspace{-0.0em}
    \begin{subfigure}[t]{0.29\textwidth}
        \centering
\includegraphics[width=\linewidth]{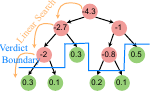}
        \caption{\bab Tree.}\label{fig:boundary}
    \end{subfigure}
    \begin{subfigure}[t]{0.29\textwidth}
        \centering
        \includegraphics[width=\linewidth]{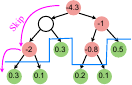}
        \caption{Our idea.}\label{fig:ouridea}
    \end{subfigure}
    \vspace{-1em}
    \caption{Verdict boundary and our idea.}
\end{wrapfigure}
\begin{myexample}\label{ex:boundary}
    Fig.~\ref{fig:boundary} depicts a \bab tree, in which the verifier assessment $\specDist$ of each node is annotated in each node. In this example, the \bab algorithm finally verifies the problem because all the subproblems are verified (i.e., $\specDist > 0$ for all the leaf nodes). We depict the verdict boundary in this \bab tree, which is $\langle 3,3,2,3,3,2 \rangle$, as illustrated by the blue line.
\end{myexample}

Recall the objective of \bab in~\S{}\ref{sec:bab}. To verify a problem, it needs to exhaustively check all the paths in a \bab tree, until it finds that every path ends with a verified leaf node (in which case, the queue $\queue$ used in Alg.~\ref{alg:bab} will be empty so it can return \true in Line~\ref{line:Qempty}). As shown in Fig.~\ref{fig:boundary}, this is equivalent to finding the verdict boundary in a \bab tree, because 
\begin{inparaenum}[1)]
    \item the verdict boundary covers all the paths;
    \item the verdict boundary signifies the shifts from unverified nodes to verified ones, and due to the monotonicity of $\specDist$ in each path, the nodes succeeding the boundary point are all verified. 
\end{inparaenum}
Therefore, for verification, the objective of \bab can be deemed to be deciding the verdict boundary in the tree.

\myparagraph{Motivation and benefits for fast boundary mining} 
We first look at how existing \bab (Alg.~\ref{alg:bab}) finds the verdict boundary. It essentially performs a \emph{linear search} in each path, i.e., it checks each node one by one along a path, as visualized in Fig.~\ref{fig:boundary}.  
Note that this way of checking is independent of how \queue in Alg.~\ref{alg:bab} works. 
Specifically, regardless of whether $\queue$ is \emph{first-in-first-out} or \emph{first-in-last-out}, for a given path, the existing \bab checks the nodes one by one.

On the other hand, due to the trend of $\specDist$ in each path, we can treat each path as a nearly sorted array, in which each element (i.e., each node) has a value $\specDist$, and consequently, the problem of deciding the  boundary point for a given path can be translated to a \emph{search problem} for the position of \emph{the first positive value} in the array. Note that, to deal with such a problem, linear search, as classic \bab does, is often known as inefficient. Instead, thanks to nearly monotonic trend of $\specDist$, we do not need to check the node one by one. For example, Fig.~\ref{fig:ouridea} shows how to locate the boundary point by skipping the second node. We can devise more efficient approaches that reduce the number of node visits, hence significantly reducing the number of calls to approximate verifiers for bounds propagation.

\vspace{-1mm}
\section{Our Proposed Approaches}
\vspace{-1mm}
Taking the insights from~\S{}\ref{sec:formulation}, in this section, we propose two novel approaches that tackle the problem of deciding the verdict boundary in a \bab tree. 

\subsection{Exponential Search Approach for Verdict Boundaries}\label{subsec:expSearch}
As mentioned in~\S{}\ref{subsec:formulation}, the \bab-based verification can be considered as a problem that aims to find the verdict boundary, and in each path of a \bab tree, finding the boundary point involves a search problem for the position of a specific value $0$ in a nearly ascending array. While existing \bab adopts linear search (see Fig.~\ref{fig:boundary}), there are some other methods, such as exponential search~\cite{bentley1976almost}, that are known to be more efficient than linear search in handling this problem. In this section, we present our proposed approach based on exponential search.

Exponential search, originally proposed in~\cite{bentley1976almost}, is an algorithm that tackles the search problem in a sorted array. Compared to other search methods, such as binary search, it can outperform if the search target is relatively close to the beginning of the array. In our case, as neural networks typically consist of a large number of neurons, as long as the position of the boundary point is not too deep in a path, this method should be suited for our purpose.

\begin{algorithm}[!tb]
\caption{Our proposed exponential search approach}
\label{alg:expSearchBaB}
\footnotesize
\begin{algorithmic}[1]
\Require A neural network $\dnn$ with $K$ neurons, an input specification $\inputSpec$, an output specification $\outputSpec$, an approximated verifier $\verifier(\cdot)$, a ReLU selection heuristic $\reluHeuristic(\cdot)$. 
\Ensure A $\mathit{verdict}\in\{\true, \false\}$.

\State $\queue\gets \{\top\}$ \Comment{initialize \queue with the original problem}
\State \Call{ExpSearchBaB}{\queue}\label{line:startExpSearch} \Comment{start the process}
\Statex 
\Function{ExpSearchBaB}{\queue}
\If{\Call{Empty}{\queue}} \label{line:QEmptyExp}
\State \Return \true \label{line:returnTrueExp} \Comment{manage to verify the problem}
\EndIf
\State $\reluSpec\gets$ \Call{Pop}{\queue} \label{line:popQExp}
\State $\istart\gets {|\reluSpec|}, \iend\gets K$ \label{line:phase1StartExp} \Comment{initialize search range $[\istart,\iend]$}
\For{$l\in \{0, 2^0,  \ldots, 2^{\lfloor \log{(K {- |\reluSpec|})} \rfloor}, K {- |\reluSpec|}\}$} \label{line:expJumpExp} \Comment{exponentially select nodes}
\State $\specDist\gets \verifier(\dnn, \inputSpec, \outputSpec, \reluSpec\land \reluAction_0 \land  \ldots  \land \reluAction_{l})$ \label{line:jumpCheckExp}
\Statex {\footnotesize \qquad\qquad where $\reluAction_0 \equiv \top$, $\reluAction_j \in \reluHeuristic(\reluSpec \land \reluAction_0 \land \ldots \land \reluAction_{j-1})$ for $j\in\{1,\ldots, l\}$}
\If{$\specDist < 0$} \Comment{the subproblem is not verified}
\State $
\begin{cases}
    \Return \;\false & \text{if } l {+ |\reluSpec|}=K \\
    \istart\gets l {+ |\reluSpec|} & \text{otherwise}
\end{cases}
$ \label{line:updateIStartP1Exp} \Comment{update \istart or return \false}
\Else
\State $\iend\gets l {+ |\reluSpec|}$ \label{line:updateIEndP1Exp} \Comment{update \iend}
\State \textbf{break} \label{line:phase1EndExp}\Comment{the range $[\istart, \iend]$ is decided}
\EndIf
\EndFor
\State $\iend \gets$ \Call{BinarySearch}{\reluSpec, \istart, \iend}
\For{$k\in \{1, \ldots, \iend {- |\reluSpec|}\}$}
\State $\queue\gets \queue\cup \{\reluSpec\land \reluAction_0 \land  \ldots \land\reluAction_{k-1}  \land \neg\reluAction_{k}\}$ 
\label{line:pushPathExp} \Comment{add unsolved paths to $\queue$}
\EndFor
\State \Return \Call{ExpSearchBaB}{$\queue$}\Comment{recursive call with updated $\queue$}
\EndFunction

\Statex 
\Function{BinarySearch}{\reluSpec, \istart, \iend}\label{line:phase2StartExp}
\While{$\iend - \istart > 1$} 
\State $m \gets \lceil\frac{\istart+\iend}{2}\rceil$ \label{line:getMidExp} \Comment{take the mid of $[\istart, \iend]$}
\State $\specDist\gets \verifier(\dnn, \inputSpec, \outputSpec, \reluSpec\land \reluAction_0\land \ldots \land \reluAction_{m {- |\reluSpec|}})$ \label{line:checkMidExp}\Comment{obtain $\specDist$}
\State $\begin{cases}
    \istart\gets m & \text{if } \specDist < 0 \\
    \iend \gets m & \text{otherwise}
\end{cases}$  \Comment{update \istart or \iend accordingly}
\State \Return \iend \label{line:phase2EndExp}
\EndWhile

\EndFunction
\end{algorithmic}
\end{algorithm}
\myparagraph{Algorithm details} Our proposed algorithm is presented in Alg.~\ref{alg:expSearchBaB}. It consists of two phases to decide the boundary point for each path: first, it identifies a range that contains the boundary point, by progressively sliding and expanding the range whose size increases exponentially; given such a range, it searches for the exact position of the  boundary point by a binary search. 

Alg.~\ref{alg:expSearchBaB} uses a queue $\queue$ to maintain the paths to be checked, which is initialized to contain $\reluSpec$ only, i.e., the original verification problem. The algorithm starts with a call of the function \textsc{ExpSearchBaB} with the initial $\queue$ (Line~\ref{line:startExpSearch}), and then in function \textsc{ExpSearchBaB}, with a ReLU specification $\reluSpec$ (i.e., a node) popped from $\queue$ (Line~\ref{line:popQExp}), it goes through the two phases of exponential search:
\begin{compactitem}[$\bullet$]
    \item The first phase (Line~\ref{line:phase1StartExp} -- Line~\ref{line:phase1EndExp}) aims to identify a range $[\istart, \iend]$ that contains the boundary point, which is initialized to be $[|\reluSpec|, K]$ in Line~\ref{line:phase1StartExp}. To achieve high efficiency, unlike existing \bab that checks the nodes sequentially, it skips nodes with an exponentially increasing step size (Line~\ref{line:expJumpExp}). For each selected node $\reluSpec\land \reluAction_0 \land \reluAction_1 \land  \ldots  \land \reluAction_{l}$, it applies \verifier to obtain $\specDist$ of the node (Line~\ref{line:jumpCheckExp}): if $\specDist < 0$ (i.e., the node is not verified), the lower bound $\istart$ of the range is updated to be the current $l$ (Line~\ref{line:updateIStartP1Exp}) and the search proceeds to the next $l$;  otherwise, the upper bound $\iend$ is updated (Line~\ref{line:updateIEndP1Exp}), and this phase can be terminated (Line~\ref{line:phase1EndExp}), because at this point, $\istart$ identifies an unverified node and $\iend$ identifies a verified node, so the boundary point must be within $[\istart, \iend]$.  
    \item The second phase (Line~\ref{line:phase2StartExp} -- Line~\ref{line:phase2EndExp}) performs a binary search to exactly find the boundary point. Here, it iteratively checks the node in the middle of $[\istart, \iend]$ (Line~\ref{line:getMidExp}, \ref{line:checkMidExp}) to obtain $\specDist$, and updates $\istart$ or $\iend$ according to the sign of $\specDist$ to further shrink the range. On the termination of this phase, the node identified by $\istart$ is unverified and the node identified by $\iend$ is verified; by Def.~\ref{def:boundary}, $\iend$ can be returned as the boundary point for this path. 
\end{compactitem}
After finding the boundary point $\iend$, Alg.~\ref{alg:expSearchBaB} pushes to $\queue$ all the unsolved paths resulted from the path $\reluSpec\land \reluAction_0 \land\reluAction_1 \land  \ldots   \land \reluAction_{\iend-|\reluSpec|}$ that has just been checked. One such unsolved path can be obtained by concatenating a prefix of the checked path with the negation of the last node, i.e., $\reluSpec\land \reluAction_0 \land\reluAction_1 \land  \ldots \land\reluAction_{k-1}  \land \neg\reluAction_{k}$, for each $k\in \{1, \ldots, \iend - |\reluSpec|\}$, in Line~\ref{line:pushPathExp}. Then, Alg.~\ref{alg:expSearchBaB} recursively calls function \textsc{ExpSearchBaB} with the updated $\queue$ to process the subsequent unsolved paths. 

Alg.~\ref{alg:expSearchBaB} can be terminated if $\queue$ is empty (Line~\ref{line:QEmptyExp}), in which case the verdict boundary for each path of the \bab tree has been found, and so Alg.~\ref{alg:expSearchBaB} can return $\true$, signifying that the problem can be verified (Line~\ref{line:returnTrueExp}).

\subsection{Gradient-Based Approach for Verdict Boundaries}\label{subsec:gradDescent}

The exponential search approach in~\S{}\ref{subsec:expSearch} allows skipping boundary-unrelated subproblems (hence avoid redundant calling to approximated verifiers) compared to linear search in classic \bab, however, it may still require a large number of visits of tree nodes before identifying the range that contains the boundary point, as it ignores the quantitative information about the change of $\specDist$ of the nodes in a path. To bridge this gap, we propose a \emph{gradient}-based search approach that leverages quantitative information about the ascending rate of nodes, allowing us to identify the verdict boundary with much fewer node visits. 

In this approach, by treating the absolute value $|\specDist|$ of $\specDist$ of each node in a path as the objective function, we can transform the problem of deciding the verdict boundary in that path, into an optimization problem that minimizes $|\specDist|$. Specifically, the optimization problem aims to identify the local minimum, namely, the position of the node for which $|\specDist| = 0$.

\begin{wrapfigure}[10]{r}{0.57\textwidth}
    \centering
    \vspace{-0.0em}
    \begin{subfigure}[t]{0.47\linewidth}
        \centering
\includegraphics[width=\linewidth]{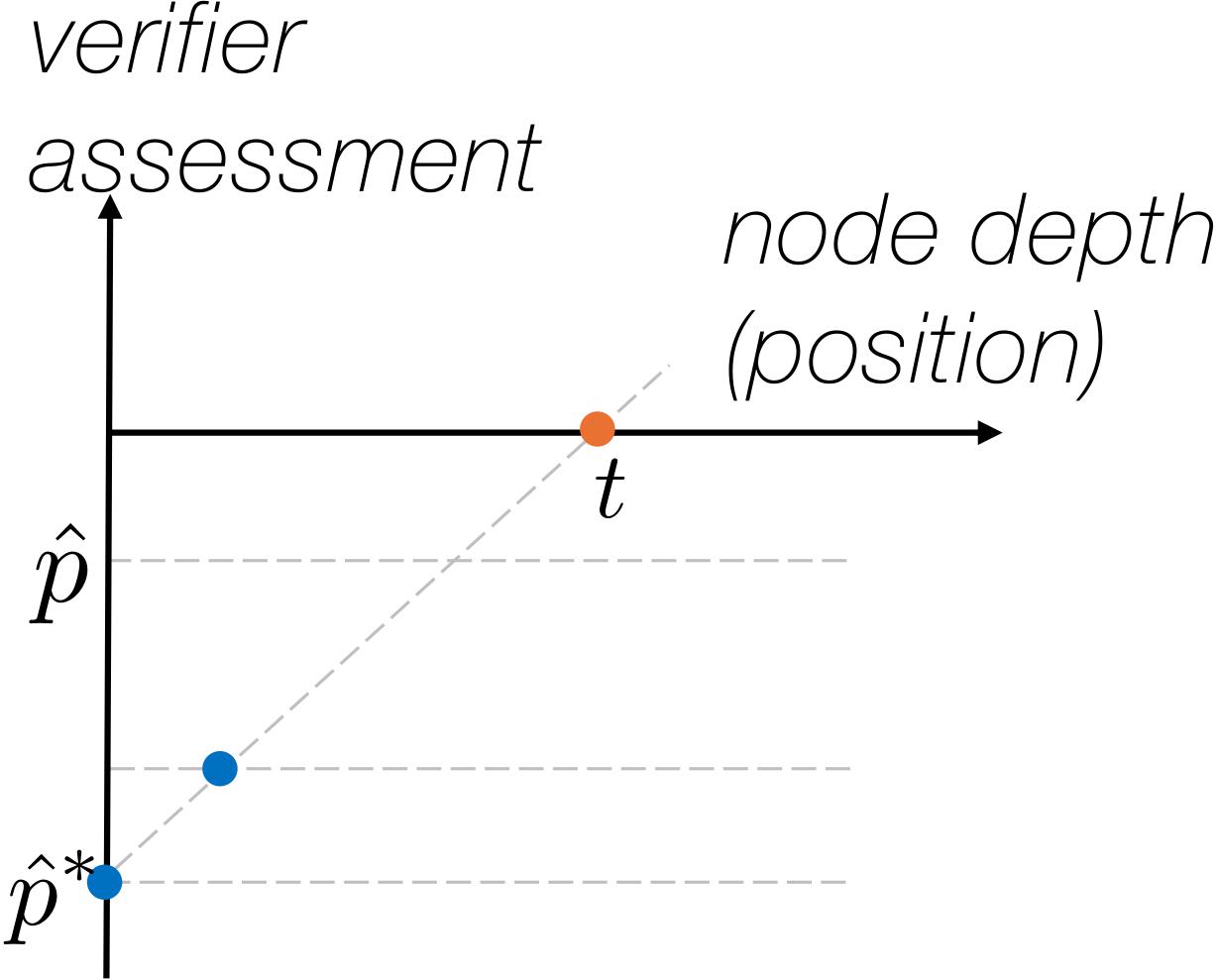}
        \caption{Estimate boundary by observed nodes.}\label{fig:under0}
    \end{subfigure}%
    \quad
    \begin{subfigure}[t][5cm]{0.47\linewidth}
        \centering
        \includegraphics[width=\linewidth]{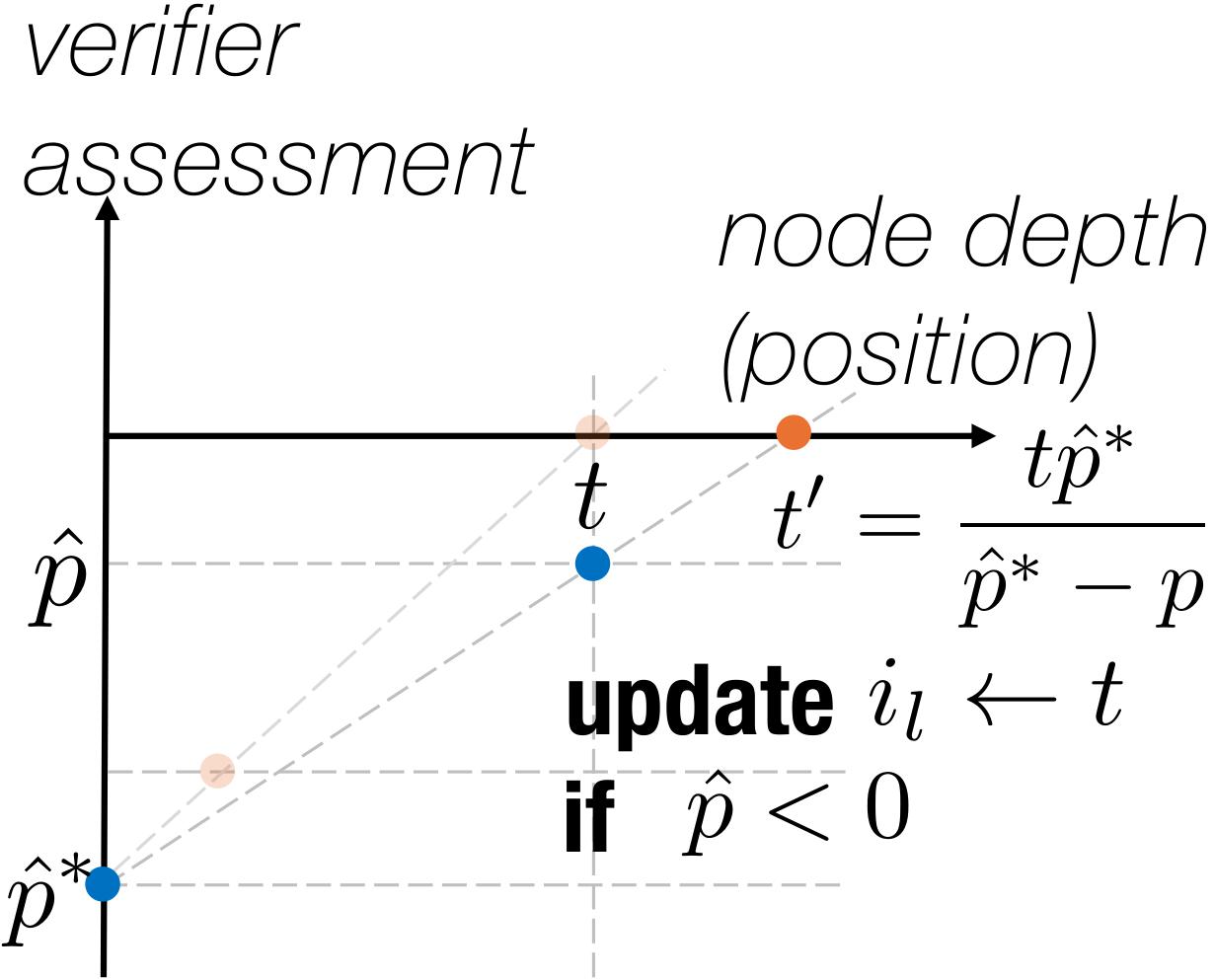}
        \caption{Update $\istart$ when $\specDist<0$ at estimated  $t$.}\label{fig:over0}
    \end{subfigure}
    \vspace{-5em}
    \caption{Gradient-based search approach.}\label{fig:gradientApproach}
\end{wrapfigure}

To solve this problem, our approach iteratively estimates the position of the local minimum, by leveraging the \emph{``gradient''} of the nodes in the path. The gradient can be estimated by exploiting the quantitative $\specDist$ of the nodes that have been visited, i.e., given two nodes $\reluSpec_1$ and $\reluSpec_2$ with verifier assessments $\specDist_1$ and $\specDist_2$ respectively, we can estimate the gradient by $\frac{\specDist_2-\specDist_1}{|\reluSpec_2|-|\reluSpec_1|}$. While this may not be precise and so our estimated local optimum can be deviated, we continue our search using the updated gradient information obtained by checking the node in the estimated position, until we find the verdict boundary.

\begin{algorithm}[!tb]
\caption{Our proposed gradient-based search approach}
\label{alg:gradDescentBaB}
\footnotesize
\begin{algorithmic}[1]
\Require A neural network $\dnn$ with $K$ neurons, an input specification $\inputSpec$, an output specification $\outputSpec$, an approximated verifier $\verifier(\cdot)$, a ReLU selection heuristic $\reluHeuristic(\cdot)$. 
\Ensure A $\mathit{verdict}\in\{\true, \false\}$.

\State $\specDist^*\gets \verifier(\dnn, \inputSpec, \outputSpec, \top)$ \label{line:checkOriginGrad} \Comment{check the original problem and record $\specDist^*$}
\If{$\specDist > 0$}
\State \Return $\true$ \Comment{verify the problem if $\specDist > 0$}
\Else
\State $\queue\gets \queue\cup \{\reluAction\}$, {for} $\reluAction\in\reluHeuristic(\top)$\label{line:initPushGrad} \Comment{add {children} to \queue}
\EndIf
\State \Call{GradSearchBaB}{\queue} \label{line:initCallGrad} \Comment{start the search process}
\Statex 
\Function{GradSearchBaB}{\queue}
\If{\Call{Empty}{\queue}}
\State \Return \true \Comment{manage to verify the problem}
\EndIf
\State $\reluSpec\gets \Call{Pop}{\queue}$
\State $\istart\gets |\reluSpec|, \iend \gets K, t\gets |\reluSpec|$ \label{line:initRangeGrad} \Comment{initialize $[\istart, \iend]$ and estimation $t$}
\State $\specDist\gets \verifier(\dnn, \inputSpec, \outputSpec, \reluSpec)$ \Comment{check the node $\reluSpec$}
\While{$\specDist < 0$} \label{line:loop1StartGrad}
\State $
\begin{cases}
    \Return \;\false & \text{if } t = K \\
    \istart\gets t \label{line:updateLowerLoop1}& \text{otherwise}
\end{cases}
$ \Comment{update \istart or return \false}
\State $t\gets \min\left(\lceil\frac{t\cdot\specDist^*}{\specDist^*-\specDist}\rceil, K\right)$\label{line:computetLoop1Grad} \Comment{estimate $t$ based on updated information}
\State $\specDist\gets \verifier(\dnn, \inputSpec, \outputSpec, \reluSpec \land \reluAction_0 \land \ldots \land \reluAction_{t-|\reluSpec|})$ \label{line:checktLoop1Grad} \Comment{check node at $t$}
\Statex {\footnotesize \qquad\qquad where $\reluAction_0 \equiv \top$,}  
\Statex {\footnotesize \qquad\qquad\qquad\;\; $\reluAction_j \in \reluHeuristic(\reluSpec \land \reluAction_0 \land \ldots \land \reluAction_{j-1})$, for $j\in\{1,\ldots, t-|\reluSpec|\}$ }
\EndWhile
\While{$\specDist > 0$}
\State $\iend\gets t$ \label{line:updateUpperLoop2} 
\Comment{update \iend if $\specDist>0$ at $t$}
\State $t\gets \max\left(\lceil\frac{t\cdot\specDist^*}{\specDist^*-\specDist}\rceil, \istart\right)$ \label{line:computetLoop2Grad} \Comment{estimate $t$ based on updated information}
\State 
$\specDist\gets \verifier(\dnn, \inputSpec, \outputSpec, \reluSpec\land \reluAction_0 \land \ldots  \land \reluAction_{t-|\reluSpec|})$\label{line:checktLoop2Grad} \Comment{check node $t$}
\EndWhile
\State $\istart\gets t$  \label{line:finalUpdateStartGrad} \Comment{update $\istart$ if $t <\istart$}
\State $\iend \gets$ \Call{BinarySearch}{\reluSpec, \istart, \iend} \label{line:phase2StartGrad}
\For{$k\in \{1, \ldots, \iend {- |\reluSpec|}\}$}
\State $\queue\gets \queue\cup \{\reluSpec\land \reluAction_0 \land  \ldots \land\reluAction_{k-1}  \land \neg\reluAction_{k}\}$ 
\label{line:pushPathGrad} \Comment{collect unsolved paths}
\EndFor
\State \Return \Call{GradSearchBaB}{$\queue$} \Comment{recursive call with updated \queue}
\EndFunction 
\end{algorithmic}
\end{algorithm}

\myparagraph{Algorithm details} Our proposed algorithm is presented in Alg.~\ref{alg:gradDescentBaB}. Similarly to exponential search, Alg.~\ref{alg:gradDescentBaB} also consists of two phases to identify the boundary point. First, it identifies a range $[\istart, \iend]$ that contains the boundary point by leveraging the gradient information; after having such a range, the second phase applies binary search to exactly locate the boundary point, similarly to Alg.~\ref{alg:expSearchBaB}.

Alg.~\ref{alg:gradDescentBaB} uses a queue $\queue$ to maintain the paths to be checked. At the beginning of Alg.~\ref{alg:gradDescentBaB}, it checks the original problem by applying $\verifier$ and obtains its verifier assessment $\specDist^*$ (Line~\ref{line:checkOriginGrad}). If this node is not verified, Alg.~\ref{alg:gradDescentBaB} pushes its children into $\queue$ (Line~\ref{line:initPushGrad}), and then calls  \textsc{GradSearchBaB} with $\queue$ to search for the verdict boundary in the \bab tree (Line~\ref{line:initCallGrad}). 

The function \textsc{GradSearchBaB} includes the two phases of searching for the boundary point in a specific path. We first elaborate on the first phase (Line~\ref{line:initRangeGrad} -- Line~\ref{line:finalUpdateStartGrad}). This phase involves two \texttt{while} loops which iteratively update the upper bound $\iend$ and the lower bound $\istart$ of the range $[\istart, \iend]$ that contains the boundary point, by iteratively checking the position where $\specDist$ is estimated to be $0$. This process is visualized in Fig.~\ref{fig:gradientApproach}, and described as follows:

\begin{compactitem}[$\bullet$]
    \item In the first loop, $t$ is the estimated position of boundary point, which is computed by taking the minimum between $\lceil\frac{t\cdot\specDist^*}{\specDist^*-\specDist}\rceil$ and $K$ (Line~\ref{line:computetLoop1Grad}), as shown in Fig.~\ref{fig:under0}. This loop applies \verifier to check the $\specDist$ of the node at $t$ (Line~\ref{line:checktLoop1Grad}), and if $\specDist$ is negative, it updates the lower bound $\istart$ to be $t$ (Line~\ref{line:updateLowerLoop1}), and re-estimates $t$ based on the updated $\specDist$ of the node {at} $t$ (Line~\ref{line:computetLoop1Grad}). The loop is terminated if $\specDist$ is positive at the estimated position, signifying that $\istart$ cannot be further refined to a greater position where $\specDist$ is still negative;
    \item The second loop is similar to the first loop, but the purpose is to refine $\iend$ to be a smaller position where $\specDist$ is still positive. It uses a similar method to estimate the boundary point position $t$, by taking the maximum between $\lceil\frac{t\cdot\specDist^*}{\specDist^*-\specDist}\rceil$ and $\istart$ (Line~\ref{line:computetLoop2Grad}), as shown in Fig.~\ref{fig:over0}. This loop applies \verifier to check the $\specDist$ of the node at $t$ (Line~\ref{line:checktLoop2Grad}), and if $\specDist$ is positive, it updates the upper bound $\iend$ to be $t$ (Line~\ref{line:updateUpperLoop2}), and re-estimates $t$ based on the updated $\specDist$ of the node {at} $t$ (Line~\ref{line:computetLoop2Grad}). The loop is terminated if $\specDist$ is negative at the estimated position, signifying that $\iend$ cannot be further refined to be a smaller position.
\end{compactitem}

Note that, during the loop, it is possible to encounter the situation that $\specDist$ does not monotonically increase as discussed in \S{}\ref{subsec:monotone}; in this case, as we show that such monotonicity often happens locally, we just skip such a non-monotonic node and use the next one. 
After the loop, if $t$ is greater than $\istart$, $\istart$ will be updated to be $t$ to further shrink the range (Line~\ref{line:finalUpdateStartGrad}). 
This range $[\istart,\iend]$ then serves as the inputs of the binary search (Line~\ref{line:phase2StartExp}) of Alg.~\ref{alg:expSearchBaB} in the second phase to exactly locate the boundary point. 

While the second phase (Line~\ref{line:phase2StartGrad}) of Alg.~\ref{alg:gradDescentBaB} also uses binary search, its input range $[\istart, \iend]$ can be much narrower than in Alg.~\ref{alg:expSearchBaB}. 
This is because the exponential search in Alg.~\ref{alg:expSearchBaB} expands the range exponentially, whereas our loops can produce a relatively smaller range.
The binary search in Alg.~\ref{alg:gradDescentBaB} can be more efficient to locate the boundary.

\vspace{-1mm}
\section{Experimental Evaluation}\label{sec:evaluation}
\vspace{-1mm}
We present the experimental setup, results, and analysis. Our implementation and data are available at \cite{fmartifact}.

\begin{wraptable}[7]{r}{0.54\textwidth}
    \centering
    \vspace{-2.2em}
    \caption{Benchmark details.}
    \label{tab:benchmark}
    \vspace{-1em}
    \resizebox{\linewidth}{!}{
    \begin{tabular}{ccccc}
        \toprule
        \#Problem & Network & Architecture & Dataset & \#ReLU \\
        \midrule
        100 & $\mnist_{\ltwo}$ & 2 $\times$ 256 linear  & \mnist & 512 \\ 
        100 & $\mnist_{\lfour}$ & 4 $\times$ 256 linear  & \mnist & 1024 \\
        \midrule
        100 & $\OVAL_{\base}$ & 2 Conv, 2 linear  & \cifar & 3172 \\  
        100 & $\OVAL_{\wide}$ & 2 Conv, 2 linear  & \cifar & 6244 \\ 
        100 & $\OVAL_{\deep}$ & 4 Conv, 2 linear & \cifar & 6756 \\ 
        \bottomrule
    \end{tabular}}
\end{wraptable}

\myparagraph{Benchmarks}
We adopt widely-used benchmarks from the neural network verification community~\cite{brix2023fourth} and five neural network models relevant to two common datasets \mnist and \cifar adopted in VNN-COMP series~\cite{brix2023fourth}.

\begin{compactitem}[$\bullet$]
\item For \mnist, we use two fully connected networks, $\mnist_{\ltwo}$ and $\mnist_{\lfour}$.
\item For \cifar, we adopt $\OVAL_{\base}$, $\OVAL_{\wide}$ and $\OVAL_{\deep}$.

\end{compactitem}

We call each pair that consists of a neural network and a specification to be a verification problem. As shown in Table~\ref{tab:benchmark}, we have a total of 500 problems that concern the five neural networks and different specifications.

\myparagraph{Baseline}
To assess  \tool,  we compare with existing tools as our baselines:
\begin{compactitem}[$\bullet$]
    \item {\it \abcrown}~\cite{wang2021beta} is the winner of VNN-COMP~\cite{brix2023fourth} that adopts classic \bab with an approximated verifier that is not monotonic under problem splitting;
    \item {\it \oliva}~\cite{zhang_ecoop} is a meta-heuristic \bab falsifier that aggressively guides the tree search toward rejecting unsafe verification instances as fast as possible.
\end{compactitem}


\myparagraph{Extension of ReLU Selection Strategy} 
Our approaches adopt the same ReLU selection strategy as baselines to ensure the fairness of comparison.
As the ReLU selection strategy is not the primary focus of this work, we make an extension of existing methods to integrate into our framework. Commonly used ReLU selection strategies, such as DeepSplit~\cite{henriksen2021deepsplit}, BaBSR~\cite{bunel2020branch} and FSB~\cite{de2021improved}, generally follow a two-step procedure: 

\begin{compactenum}[(1)]
    \item each neuron is assigned a score based on its potential abstraction–refinement gain and its estimated impact on the final network output
    \item the neuron with the highest score (Top-1) is selected for splitting.
\end{compactenum} 

In our approach, we adopt and extend the state-of-the-art ReLU selection strategy~\cite{de2021improved}. While Step (1) remains unchanged, we extend Step (2) by selecting the Top-\emph{k} neurons to split simultaneously, where $k$ corresponds to the number of neurons skipped by our method. This extension allows our verification approach to retain compatibility with existing selection heuristics while improving efficiency through parallel refinement of multiple high-impact neurons. 

This extension further inspires the development of novel ReLU selection heuristics incorporating abstraction–refinement prediction mechanisms, which could better complement our \tool. As designing such strategies is non-trivial, we leave this as future work.

\myparagraph{Evaluation metrics} In our experiments, to compare the performances of different approaches, we apply both the baseline approaches and our proposed approaches \toolb, \toolg, to all the verification problems. 
For each problem, we set 1,000 seconds as the timeout. We adopt the following metrics:
\begin{compactitem}[$\bullet$]
    \item \emph{Verification result}: It indicates whether an approach can solve a problem (i.e., it returns either \true or \false for the problem) within the given time budget;
    \item \emph{Time cost}: In this case, if an approach successfully solves a problem, we record the time taken as an indicator of its efficiency.
\end{compactitem}
\myparagraph{Experiment environment}
The experiments were conducted on
AWS EC2 instance with 32GB memory and 8-core CPUs.

\vspace{-1em}
\subsection{Evaluation Results}
\vspace{-1em}

\researchquestion{Are our proposed approaches more efficient than the baseline approach?}
\begin{wraptable}[9]{r}{0.5\textwidth} 
    \centering
    \footnotesize
    \vspace{-2.7em}
    \caption{RQ1 -- Performance comparison on certified (C) and falsified (F) counts.}
    \label{tab:overall}
    \resizebox{0.5\textwidth}{!}{%
    \begin{tabular}{l|p{2em}p{2em}|p{2em}p{1.5em}|p{2em}p{1.5em}|p{2em}p{1.5em}}
    \toprule
        Model   & \multicolumn{2}{c}{\oliva} & \multicolumn{2}{c}{\abcrown} & 
        \multicolumn{2}{c}{\toolb} & \multicolumn{2}{c}{\toolg} \\ 
         & C & F  & C & F  & C &F &C& F \\\hline
$\mnist_{{{\ltwo}}}$ & 72 & 7&89 &4 &89 &4&89& \tbgreen 7 \\
$\mnist_{{{\lfour}}}$&39& 10 &51 &2&51 & 2 &51&2\\
$\OVAL_{{\base}}$&69& 10 &83&0 &\tbgreen 85&1&\tbgreen 85&1\\
$\OVAL_{{\deep}}$ &53& 8&71&0&71&0 &\tbgreen 76& 1\\
$\OVAL_{{\wide}}$ &51& 9&70&1&70&1 &\tbgreen 72&1
\\\bottomrule
    \end{tabular}%
    }
\end{wraptable}
\myparagraph{Comparison with baselines} 
Table~\ref{tab:overall} reports the number of verification instances. For certified instances (C), \abcrown serves as a strong baseline, certifying more instances than \oliva. Our approaches, \toolb and \toolg, consistently surpass \oliva, certifying more instances than it, while matching or slightly exceeding \abcrown. This demonstrates the strengths of our approaches. On the other hand, \oliva maintains an advantage in falsification (F), consistently identifying more counterexamples due to its design, which focuses on rapid falsification rather than exhaustive proof.

\myparagraph{Efficiency comparison relative to \abcrown} 
To further highlight the benefits of our approaches, Table~\ref{tab:efficiency1} reports the relative change in verification time compared to \abcrown. Notably, \toolg achieves substantial time savings, with average reductions of 17.0\%--29.2\% and up to 44.7\% in best-case scenarios, thanks to the gradient-based estimation strategy that skips redundant computations. \toolb provides a smaller but consistent average improvement of about 4\%. In worst-case scenarios, the aggressive search strategy can introduce overhead (up to 16.7\%), caused by backtracking when the initial estimation overshoots the verdict boundary (Lines~\ref{line:updateUpperLoop2}--\ref{line:checktLoop2Grad} in Alg.~\ref{alg:gradDescentBaB}). Overall, these results demonstrate that, compared to \abcrown, our approaches bring an evident improvement in terms of verification efficiency.

\begin{table}[!t]
    \centering
    \footnotesize
    \caption{RQ1 -- Average, best, and worst-case time comparison w.r.t. \abcrown.
    }
    \label{tab:efficiency1}
    \resizebox{0.55\textwidth}{!}{%
    \begin{tabular}{l|ccc|ccc }
    \toprule
        Model & 
        \multicolumn{3}{c}{\toolb} & \multicolumn{3}{c}{\toolg} \\ 
      &AVG & BC&WC & AVG& BC& WC\\\hline
$\mnist_{{{\ltwo}}}$ &  -4.5\% &-12.9\%& +5.6\%&\tbgreen -29.2\% &\tbgreen -44.7\% &+13.0\% \\
$\mnist_{{{\lfour}}}$& -4.7\% &-11.5\%&+5.1\%&\tbgreen  -17.9\% & -34.2\%&\tbred +16.7\%\\
$\OVAL_{{\base}}$& -4.0\% &-13.8\%&+4.1\%&\tbgreen -17.9\% & -35.8\% & +8.4\%\\
$\OVAL_{{\deep}}$ & -3.9\% &-13.2\%&+5.8\%&\tbgreen -18.1\% & -34.5\% & +8.4\%\\
$\OVAL_{{\wide}}$ & -4.5\% &-10.5\%&+3.0\%&\tbgreen -17.0\% & -26.9\%&+5.8\%
\\\bottomrule
    \end{tabular}
    }
     \end{table}
\myparagraph{Impact of non-monotonicity} Since non-monotonicity does not affect the soundness of our approach, we mainly demonstrate our hypothesis that it does not hinder our performance. We present an analysis of the performance comparison of our approaches for monotonic paths and non-monotonic paths, by which we can see that the impact of non-monotonicity is not significant. 

\begin{figure}[htpb]
    \centering
    \begin{subfigure}[t]{0.3\linewidth}
        \centering
\includegraphics[width=\linewidth]{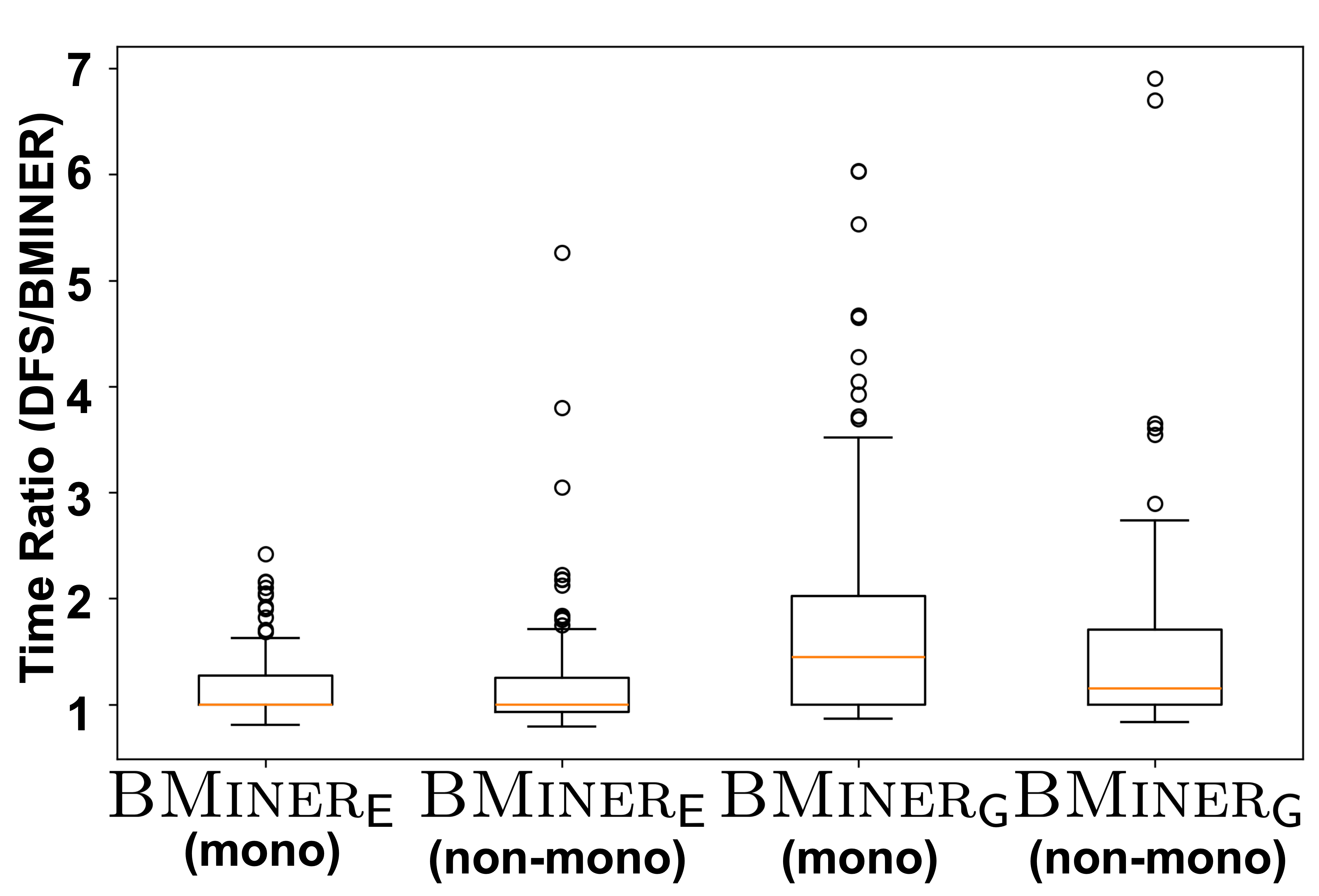}
\vspace{-1.5em}
\captionsetup{font=f7,labelfont=f7}
\caption{$\mnist_{\lfour}$}\label{fig:L4}
    \end{subfigure}
    \begin{subfigure}[t]{0.3\linewidth}
        \centering
        \includegraphics[width= \linewidth]{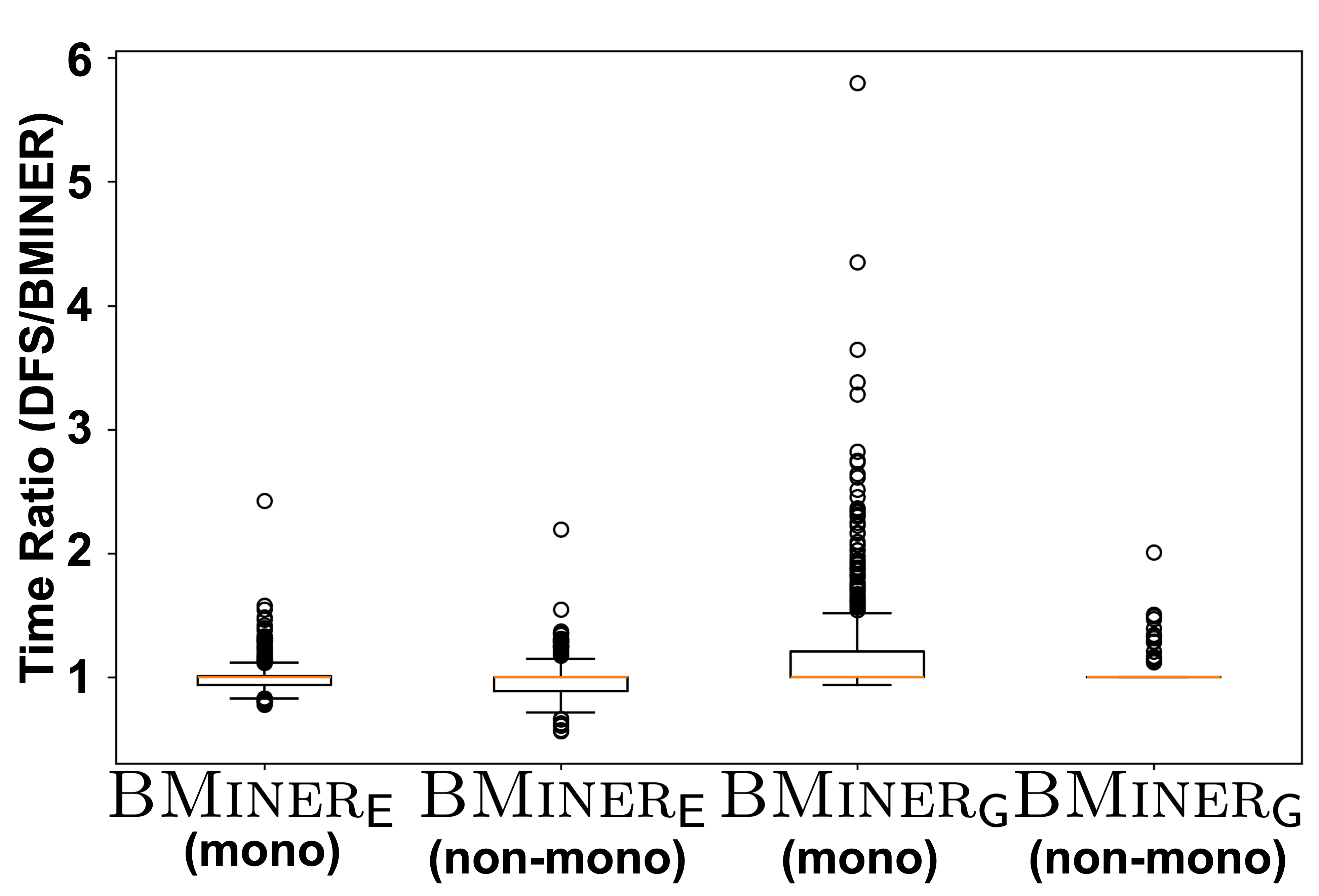}
        \vspace{-1.5em}\captionsetup{font=f7,labelfont=f7}\caption{$\OVAL_{\base}$}\label{fig:base}
    \end{subfigure}

    \begin{subfigure}[t]{0.3\linewidth}
        \centering
\includegraphics[width= \linewidth]{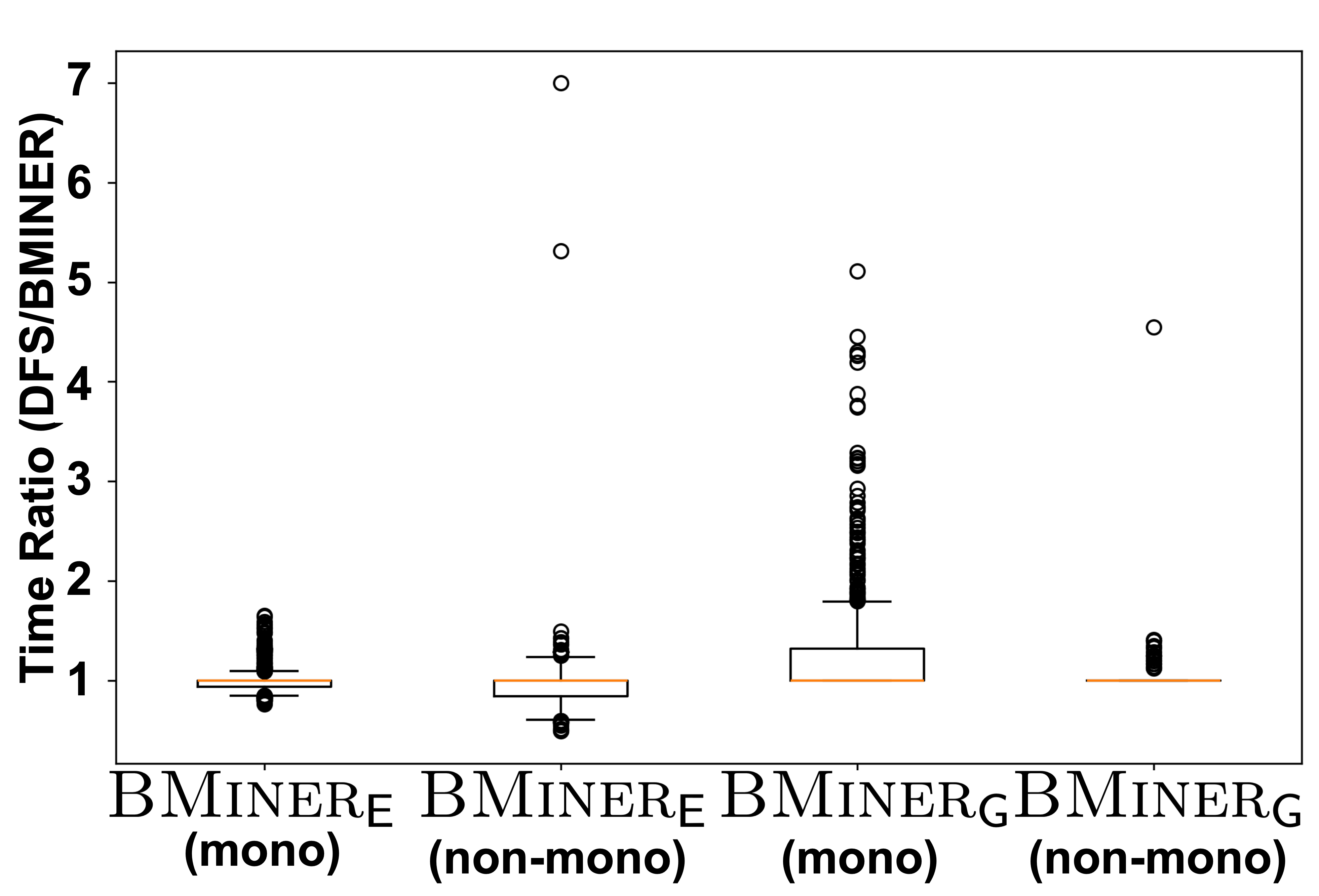}
        \vspace{-1.5em}\captionsetup{font=f7,labelfont=f7}\caption{$\OVAL_{\wide}$}\label{fig:wide}
    \end{subfigure}
\begin{subfigure}[t]{0.3\linewidth}
        \centering
        \includegraphics[width= \linewidth]{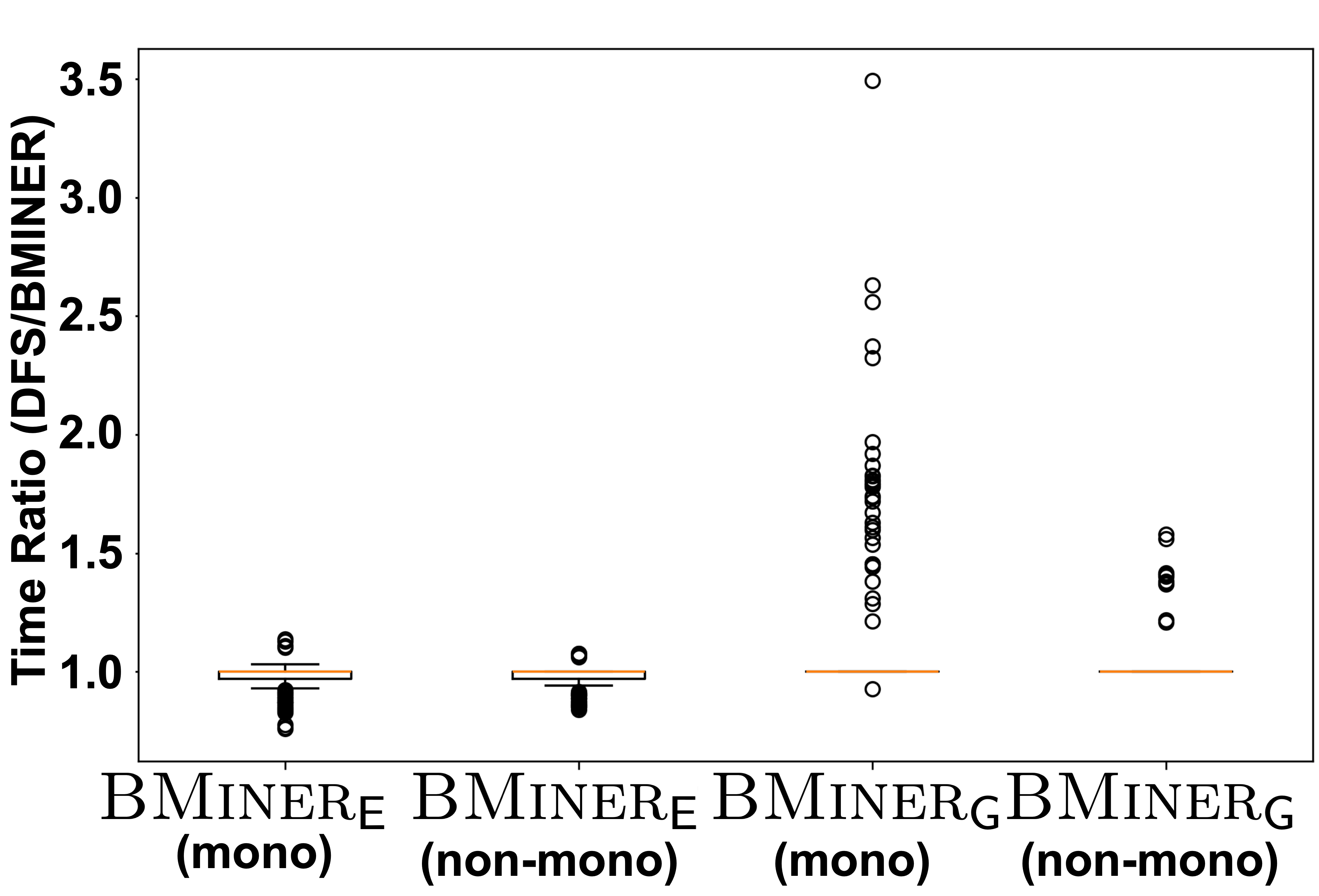}
        \vspace{-1.5em}\captionsetup{font=f7,labelfont=f7}\caption{$\OVAL_{\deep}$}\label{fig:deep}
    \end{subfigure}
    \vspace{-3mm}
    \caption{Comparison of time ratios between monotonic and non-monotonic paths for \toolb and \toolg. }
    \label{fig:impact_non_mono}
\end{figure}

To evaluate the impact of non-monotonicity on the efficiency of our verification strategies, we compare the runtime performance between monotonic and non-monotonic paths across different benchmark networks. 
We exclude $\mnist_{\ltwo}$ from the results because it does not generate a meaningful number of non-monotonic paths.
Specifically, we recorded the execution time and monotonicity status of all analyzed paths. For both monotonic and non-monotonic paths, we computed the time reduction ratio, defined as the running time of linear search divided by the running time of our proposed methods \toolb and \toolg, respectively. We further conduct experiments on $\mnist$ and $\OVAL$ models. 

The distribution of time ratio (for comparison, we compared each approach with a naive baseline that explores a \bab tree by depth first search (DFS)) in Fig.~\ref{fig:impact_non_mono} shows the consistency between monotonic and non-monotonic paths:
\begin{compactitem}
    \item \textbf{Analysis of \toolb:} Comparing on $\mnist_{\lfour}$ and $\OVAL_{\base}$ in Fig.~\ref{fig:L4} and Fig.~\ref{fig:base}, both monotonic and non-monotonic groups exhibit a similar median time ratio, indicating that \toolb achieves consistent speedup over DFS regardless of path monotonicity. The interquartile range and outliers are slightly smaller in the non-monotonic group, further supporting the robustness of \toolb in handling non-monotonic paths.

    \item \textbf{Analysis of \toolg:} Comparing on $\OVAL_{\wide}$ and $\OVAL_{\deep}$ in Fig.~\ref{fig:wide} and Fig.~\ref{fig:deep}, the median time ratio for both groups is around 1.0, indicating that \toolg consistently improves efficiency over DFS across both path types. However, the monotonic group exhibits a wider interquartile range and more high-value outliers (some exceeding 5), whereas the non-monotonic group shows a narrower spread with almost no extreme outliers. This suggests that non-monotonicity does not hinder—and may even contribute to—more stable and predictable performance under \toolg.
\end{compactitem}
In conclusion, this supports our hypothesis that the non-monotonicity of certain paths has little impact on the effectiveness of our techniques.

\begin{figure}[t]
    \centering
    \includegraphics[width=\linewidth]{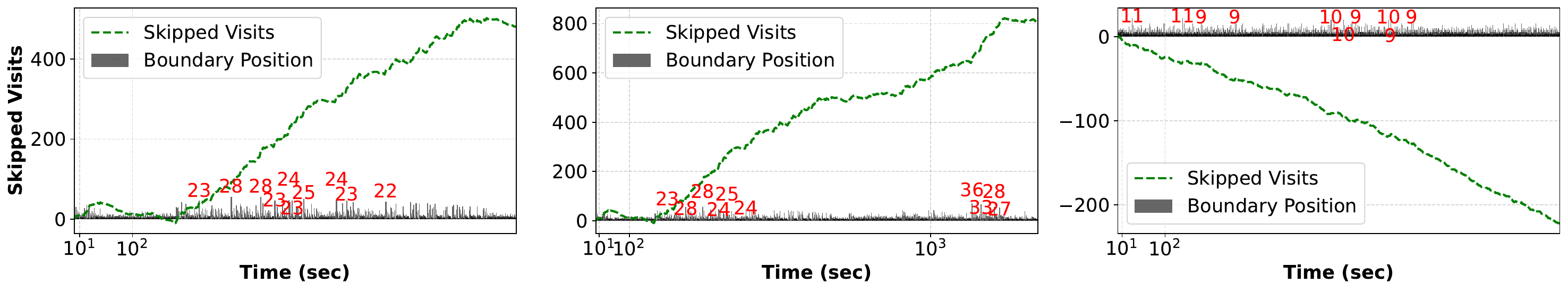}
    \vspace{-3mm}
    \caption{
RQ2 -- Our performance advantage throughout verification processes.
}
    \vspace{-3mm}
    \label{fig:evolution}
\end{figure}

\researchquestion{Which factors affect the performance of our approaches?}
We look into the verification processes of our two approaches to understand the reasons behind performance advantages.  
To that end, we select 3 problems, and show how our performance advantages (in terms of the number of skipped node visits, compared to \abcrown) evolve throughout a verification process. 

Our experimental results are shown in Fig.~\ref{fig:evolution}. In each sub-figure, $y$-axis denotes the number of skipped nodes, compared to the hypothetical number of visits by \abcrown for covering the same tree nodes. 
By the results, in the problems where \tool outperforms \abcrown, the number of skipped node visits keeps increasing in most cases. This result implies that, our strategies are indeed helpful to save subproblem solving/visits in verification. 

Moreover, we also annotate the position of a boundary point in each path, i.e., the search goal of our approaches. 
By observing this, we find that our performance advantages stem from the ability of our approach in handling long paths, because when such a long path occurs, the improvement rate will often increase. 
Over the verification process, there are many paths where the boundary point is small, for which the naive linear search in \bab-baseline can also work well; for this reason, our improvement rate can decrease as the verification process evolves. 
This can also explain when our approaches cannot perform well, as in the last sub-figure: it happens when there is almost no long path in the \bab tree.

\researchquestion{What causes performance differences between \toolb and \toolg?}\label{rq3}
In this RQ, we compare the search strategies used in \toolb and \toolg to locate the boundary point in a path. We select two verification problems for which \toolg outperforms \toolb, and we look into the number of node visits consumed by \toolb and \toolg for each path during each verification process, to understand the performance advantage of \toolg.

Table~\ref{tab:visits} shows the total number of visits for paths with different lengths. We find that \toolg has advantages in handling long paths. For example, in the first problem, for paths with length 16-40, \toolg takes an average of 6.3 visits to locate the boundary, while \toolb requires 11.2 visits. These results demonstrate the usefulness of our estimation strategy in \toolg, and explain why \toolg can outperform \toolb.
Moreover, by observing the detailed node visits for Phase 1 and Phase 2 in each approach, \toolb in Phase 1 can take fewer visits for shorter paths than \toolg, but it takes more visits for longer paths. 
In Phase 2, \toolg generally requires fewer visits than \toolb, particularly on long paths (e.g., 16–40 in the second problem), as our effective boundary estimation yields a much narrower search range.

\begin{table}[!tb]
    \centering
    \footnotesize
    \caption{RQ3 -- Node visits by \toolb and \toolg. \# denotes the number of paths with a specific length. Each value shows the average number of visits.}\label{tab:visits}
    {\fontsize{6}{6}
        \begin{tabular}{cc|ccc|ccc}
            \toprule
            \multirow{2}{*}{len.} & \multirow{2}{*}{\#}  & \multicolumn{3}{c|}{\toolb} & \multicolumn{3}{c}{\toolg} \\
            & & Ph1 & Ph2 & Total & Ph1 & Ph2 & Total \\\midrule
            
            0-7 & 70  & 4.2 & 1.4 & 5.6 & 5.2 & 1.9 & 7.1 \\  
            8-15 & 26 & 6.0 & 3.0 & 9.0 & 4.5 & 2.3 & 6.8\\ 
            16-40 & 8 &  7.1&  4.1 & \tbgreen 11.2 & 3.3 & 3.0 & \tbgreen 6.3\\
            \bottomrule
        \end{tabular}
        \begin{tabular}{cc|ccc|ccc}
            \toprule
            \multirow{2}{*}{len.} & \multirow{2}{*}{\#}  & \multicolumn{3}{c|}{\toolb} & \multicolumn{3}{c}{\toolg}\\
            & & Ph1 & Ph2 & Total & Ph1 & Ph2 & Total \\
            \midrule
            0-7 & 1151 & 3.2 & 0.5 & 3.7 & 5.5 & 0.6 & 6.1 \\ 
            8-15 & 441 & 5.9& 2.9 & 8.8 & 6.6 & 0.8 & 7.4\\ 
            16-40 & 367 & 7.0& \tbgreen 4.0& 11.0 & 5.3 & \tbgreen 0.8 & 6.1\\
            \bottomrule
        \end{tabular}}
\vspace{-2mm}
\end{table}

\vspace{-2mm}
\section{Related Work}
\vspace{-2mm}
Neural network verification has been extensively studied~\cite{ijcai2022p503,yedifm24,juncav24,katz2017reluplex,ehlers2017formal,huang2017safety,singh2018deepz,singh2019abstract, shi2022efficiently, liu2021algorithms, boudardara2024review, wu2024marabou,dalrymple2024towards,huang2024towards,wu2024verix,strong2023global,isac2023dnn,wei2023convex,wu2023toward,ostrovsky2022abstraction,zelazny2022optimizing,mitra2024formal}. 
Notably, the combined use of  \bab~\cite{bunel2020branch} with approximated methods is both efficient and complete, and has been adopted by many advanced tools such as \abcrown~\cite{wang2021beta}.
There has been a rich body of literature~\cite{henriksen2021deepsplit,de2021improved,geng2023towards,zhang2020verification,cohen2022tighter} that aims to improve \bab, but mostly they focus on issues such as splitting strategy~\cite{wang2018formal}, selection of ReLU functions~\cite{henriksen2021deepsplit,de2021improved}, counterexample-guided falsification~\cite {zhang_ecoop,guo2021eager, kota_date}.
In contrast to the classical one-by-one processing, our contribution is a new per-path formulation that jointly splits multiple activation functions to speed up BaB-based certification-centric verification.

\vspace{-2mm}
\section{Conclusion and Future Work}
\vspace{-2mm}
This paper introduces \tool, which reformulates \bab-based verification as verdict boundary mining via exponential and gradient-based searches.
\tool exploits the near-monotonic nature of verification assessments to skip redundant subproblems computations, which outperforms state-of-the-art verification approaches and reduces the average verification time by 17\% to 30\%, with a maximum time reduction of 44.7\%. As future work, we aim to further improve the efficiency by considering GPU parallelization, meta-heuristic falsification, and the order of checking different paths in verdict boundary mining, which can affect the efficiency of our approaches.

%
%
%
\vspace{-1mm}
\section*{Acknowledgements}
\vspace{-1mm}
We thank the anonymous reviewers for their comments and suggestions. This research is supported by the Australian Research Grant DP250101396, the JST BOOST Grant No. JPMJBY24D7 and the JSPS Grant No. JP25K21179.

\bibliographystyle{splncs04}
\bibliography{fm2026}

\clearpage

\end{document}